\documentclass{article}

\usepackage[final]{neurips_2023}

\usepackage[utf8]{inputenc} 
\usepackage[T1]{fontenc}    
\usepackage{hyperref}       
\usepackage{url}            
\usepackage{booktabs}       
\usepackage{amsfonts}       
\usepackage{nicefrac}       
\usepackage{microtype}      
\usepackage{xcolor}         
\usepackage{geometry}
\usepackage{pdflscape}
\usepackage{adjustbox}
\usepackage{tcolorbox}
\usepackage{graphicx} 
\usepackage{makecell, booktabs}
\usepackage{inconsolata} 
\usepackage{bold-extra} 

\title{Multimodal Auto Validation For Self-Refinement in Web Agents}

\author{%
  Ruhana Azam\\
  University of Illinois\\
  \texttt{razam2@illinois.edu} \\
  \And
  Tamer Abuelsaad\\
  Emergence.ai\\
  \texttt{tea@emergence.ai} \\
  \And
  Aditya Vempaty\\
  Emergence.ai\\
  \texttt{aditya@emergence.ai} \\
  \And
  Ashish Jagmohan\\
  Emergence.ai\\
  \texttt{ashish@emergence.ai} \\
}
\begin{document}

\maketitle

\begin{abstract}
As our world digitizes, web agents that can automate complex and monotonous tasks are becoming essential in streamlining workflows. This paper introduces an approach to improving web agent performance through multi-modal validation and self-refinement. We present a comprehensive study of different modalities (text, vision) and the effect of hierarchy for the automatic validation of web agents, building upon the state-of-the-art Agent-E web automation framework. We also introduce a self-refinement mechanism for web automation, using the developed auto-validator, that enables web agents to detect and self-correct workflow failures. Our results show significant gains on Agent-E's (a SOTA web agent) prior state-of-art performance, boosting task-completion rates from 76.2\% to 81.24\% on the subset of the WebVoyager benchmark. The approach presented in this paper paves the way for more reliable digital assistants in complex, real-world scenarios.
\end{abstract}

\section{Introduction} 
    Recent studies indicate that an estimated $92\%$ of jobs now require digital skills~\citep{bergson2023closing}, driving companies to focus on workflow automation. Generative AI and other automation tools are particularly promising, with the potential to handle 60-70\% of an employee's tasks and contribute between \$2.6 trillion and \$4.4 trillion to global GDP~\citep{chui2023economic}. In this landscape, web agents can play an essential role in alleviating workloads. Acting as personal assistants, they can manage schedules, automate routine activities, and efficiently complete simple tasks. By reducing cognitive load, saving time, and optimizing workflows, these tools help individuals and businesses operate more efficiently.

Earlier web agents were primarily rule-based, which limited their adaptability across diverse scenarios. However, advances in Large Language Models (LLMs), known for achieving human-like performance in various tasks~\citep{liang2022holistic}, have sparked renewed interest in leveraging LLMs for web automation. Initial successes have been reported, with LLM-based approaches achieving success rates of 75\% in WebArena benchmarks~\citep{pan2024autonomous} and 73.1\% in WebVoyager~\citep{Abuelsaad2024}. While these results are promising, they highlight the need for further improvements to meet the reliability standards required for widespread adoption.

We propose using a self-refinement method~\citep{Madaan2023} to enhance web navigation agents. This method employs a validator to provide feedback and self-correct workflow failures. Like many real-life planning settings, web- navigation, in non-synthetic settings, does not have an inherent reward model. Drawing inspiration from LLM-as-a-judge~\citep{zheng2024judging}, we aim to build a task-agnostic validation system for web tasks without human supervision. Our validator achieves over 84\% agreement with human annotators on WebVoyager tasks, and we demonstrate that the generated labels can improve the performance of web navigation planners. By utilizing feedback from the validator for unsuccessful workflows, our self-refinement mechanism improves agent performance on a subset of WebVoyager tasks from 76.2\% to 81.24\%.:

Our main contributions are:
\begin{itemize}
    \item Providing a comprehensive understanding of how different modalities can be utilized to build a reliable task-agnostic validator for web workflows. This includes understanding the effect of using a hierarchy of agents to distill information that is fed into the validator. 
    \item Showing that self-refinement can reach SOTA performance for web agents in the Web Voyager tasks without any additional supervision from humans. This proves that agents can self-refine for web navigation and the benefits can be seen with self-validation. 
    \item Highlighting some practical issues that may be faced when implementing a web agent. These findings underscore the complexity of real-world web environments and provide valuable insights for future research.
\end{itemize}

Moreover, unlike prior work~\citep{Putta2024, Lutz2024, Koh2024, zhou2023webarena}, which primarily evaluates synthetic settings, our approach focuses on real-world applications using the Web Voyager benchmark~\citep{he2024webvoyagerbuildingendtoendweb}. This shift toward real-world evaluation not only showcases the adaptability of our approach but also underscores the practical challenges of deploying web agents in live environments.
\section{Background} 
    \paragraph{Self-Improving Agents}
Recent research has focused on enhancing the capabilities of LLMs during training or inference without additional human supervision~\citep{wei2022chain, chen2023universal, wang2023selfconsistencyimproveschainthought, kojima2023largelanguagemodelszeroshot}. Techniques like chain-of-thought prompting and self-consistency, as used in ~\citet{Huang2022}, aim to generate higher-quality outputs. Other methods, such as Self-refine~\citep{Madaan2023}, Reflexion~\citep{shinn2023reflexionlanguageagentsverbal}, and REFINER~\citep{paul-etal-2024-refiner}, focus on iterative refinement of outputs using actor-critic pipelines or task decomposition. These approaches have been successfully applied to web agents, improving the performance of LLMs in web automation tasks~\citep{Putta2024, pan2024autonomous, Lutz2024}.

\paragraph{Web Agents}
The reasoning and planning capabilities in large language models (LLMs) have sparked interest in developing autonomous web agents capable of navigating complex online environments. Several frameworks have emerged to enhance LLM performance in agentic settings, such as chain-of-thought prompting \citep{wei2022chain} and the ReAct paradigm \citep{yao2022react}, which combine reasoning and action steps. Leveraging such paradigms has shown promising results in prompting web agents \citep{he2024webvoyagerbuildingendtoendweb}. Moreover, using these paradigms with multimodal observations has been shown to boost performance \citep{Koh2024} further.

Researchers have also explored the use of more traditional search procedures in conjunction with self-critique mechanisms. \citet{Putta2024} has used self-critique as an intermediate reward signal for Monte Carlo Tree Search (MCTS) \citep{Putta2024}. A similar approach by \citet{Lutz2024} includes a planner that backtracks using self-critique feedback. These labeled trajectories are then used in context to plan future tasks. Other work, inspired by Reflexion~\citep{shinn2023reflexionlanguageagentsverbal}, has focused on providing reward signals based on visual information at the end of each workflow \citep{pan2024autonomous}.

Websites are textually represented with Document Object Models (DOMS). While most works have focused on the use of raw DOM and/or screenshot signals for planning \citep{he2024webvoyagerbuildingendtoendweb, pan2024autonomous, Putta2024, Lutz2024}, a different line of work has focused on the effect of making DOMs more interpretable for LLM-based planners. \citet{Abuelsaad2024} utilizes a hierarchy of agents to create a more interpretable understanding of the DOM and sequence of actions for the planner agent.

\paragraph{Web Navigation Benchmarks}
As the capabilities of digital web agents have improved, benchmarking methods have become increasingly sophisticated. While early research relied on simplified website simulations for controlled studies \citep{shi2017world, liu2018reinforcement}, recent efforts have developed more realistic environments that emulate or locally host dynamic websites \citep{deng2023mind2web, zhou2023webarena, yao2022webshop}. However, these environments still lack certain unpredictable elements found in real-world websites. Addressing this limitation, WebVoyager provides a benchmark comprising 643 tasks across 15 real-world websites \citep{he2024webvoyagerbuildingendtoendweb}. Current state-of-the-art performance on the WebVoyager dataset stands at $73.1\%$ \citep{Abuelsaad2024}. The development and evaluation of reliable web agents remain active areas for improvement.

\section{Self-Refinement For Web-Navigation} 
\begin{figure}[h!]
    \centering
    \includegraphics[width=\linewidth]{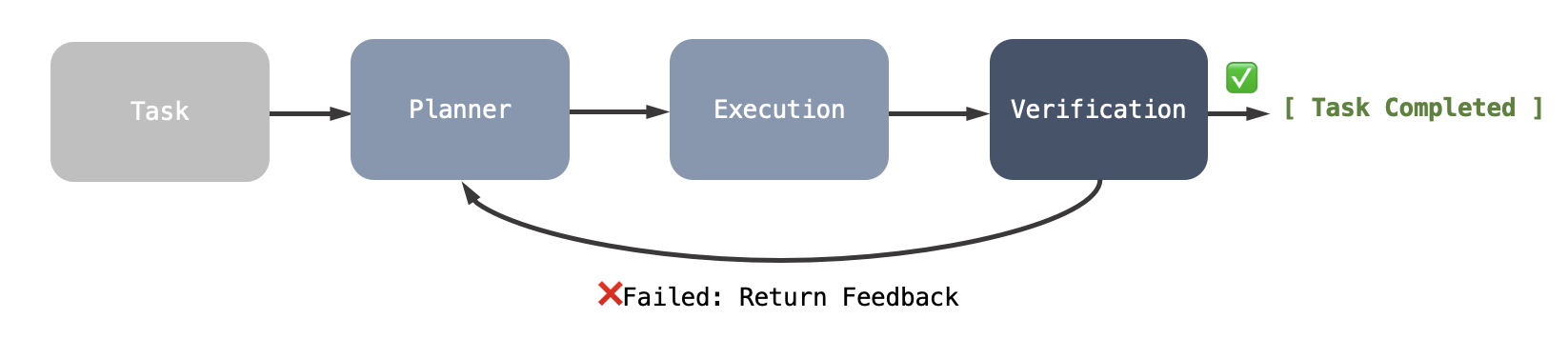}
    \caption{Self-refinement workflow.}
    \label{fig:self_refinement}
\end{figure}

    In this paper, we propose a self-refinement mechanism for web navigation tasks. A web navigation agent plans and executes a workflow for a given task, after which a validator assesses whether the task was successfully completed and provides feedback. If the task is incomplete, the agent revises its plan based on the feedback and reattempts the task. This mechanism is illustrated in Figure~\ref{fig:self_refinement}.

In Section~\ref{sec:problem_setting}, we specify our problem setting and explain how the task of web navigation can be viewed as a planning problem. Before implementing the self-refinement method, we need to build a validator agent. In Section~\ref{sec: val_method}, we introduce our methods for building the validator. Finally, in Section~\ref{sec:self_refine_method}, we discuss our implementation of the self-refinement method.

    \subsection{Problem Setting} \label{sec:problem_setting}
The web can be formalized as a Markov Decision Process (MDP) defined by the tuple $(S, A, P, R, \gamma)$, where $S$ represents web pages, $A$ denotes user actions (e.g., link clicks), $P$ comprises mostly deterministic state transitions, $R$ quantifies reward function, and $\gamma$ models exploration depth. This formulation frames web navigation as a planning problem: given $M = (S, A, R, P, \gamma)$, find the policy $\pi^*$ that maximizes expected cumulative reward $\mathbb{E}\left[\sum_{t=0}^{\infty} \gamma^t R(s_t, a_t)\right]$. Although the state and action space is known in such a setting, the transition probability and the reward model are unknown. To this end, in Section~\ref{sec: val_method},  we attempt to build a validation model that can act as a reward signal in any web navigation environment. 

\subsection{Validation Methods} \label{sec: val_method}
In this section, we describe our method for building a domain-agnostic web validator, which can act as a reward signal. Building on the concepts of LLM-as-a-judge~\citep{zheng2024judging} and self-critique mechanisms, we utilize (V)LLMs for validation. Moreover, given that the (V)LLMs are highly sensitive to their context windows, we investigate the use of text, vision, multimodality, and hierarchical task summarization for an LLM-based validator.
 
Prior work has suggested that providing multimodal observations leads to the best performance in LLM-based planners \citep{Koh2024, he2024webvoyagerbuildingendtoendweb}. Other research has pointed out that DOM observations are difficult for an LLM-based planner to interpret due to their size and verbosity, often containing irrelevant information for the task at hand. While different methods of prompting planner agents have been tested, the best practice for validation remains unclear. To address this, we investigate the use of text, vision, multimodality, and hierarchical task summarization for an LLM-based validator.

\begin{figure}[h]
    \centering
    \includegraphics[width=0.7\linewidth]{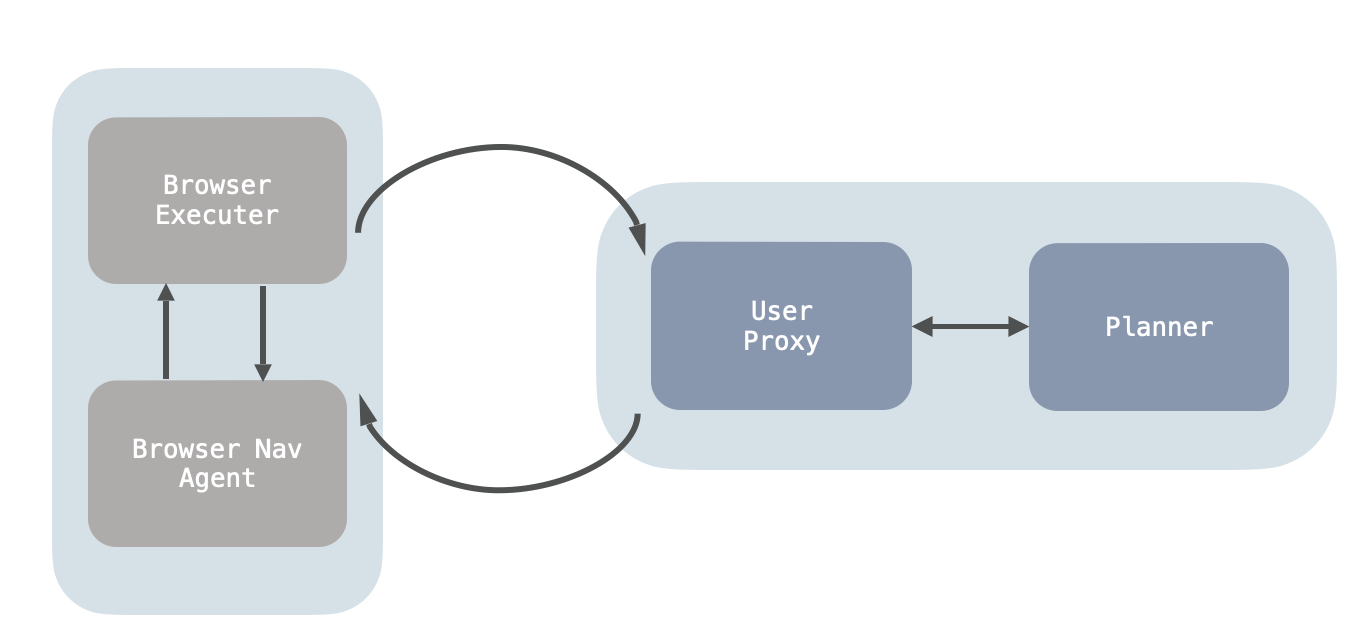}
    \caption{Multi-Agent Architecture in Agent-E.}
    \label{fig:agent_e}
\end{figure}

\begin{figure}[h!]
    \centering
    \setlength{\fboxsep}{0pt}
    \setlength{\fboxrule}{0.5pt}
    \fbox{\includegraphics[width=0.98\linewidth]{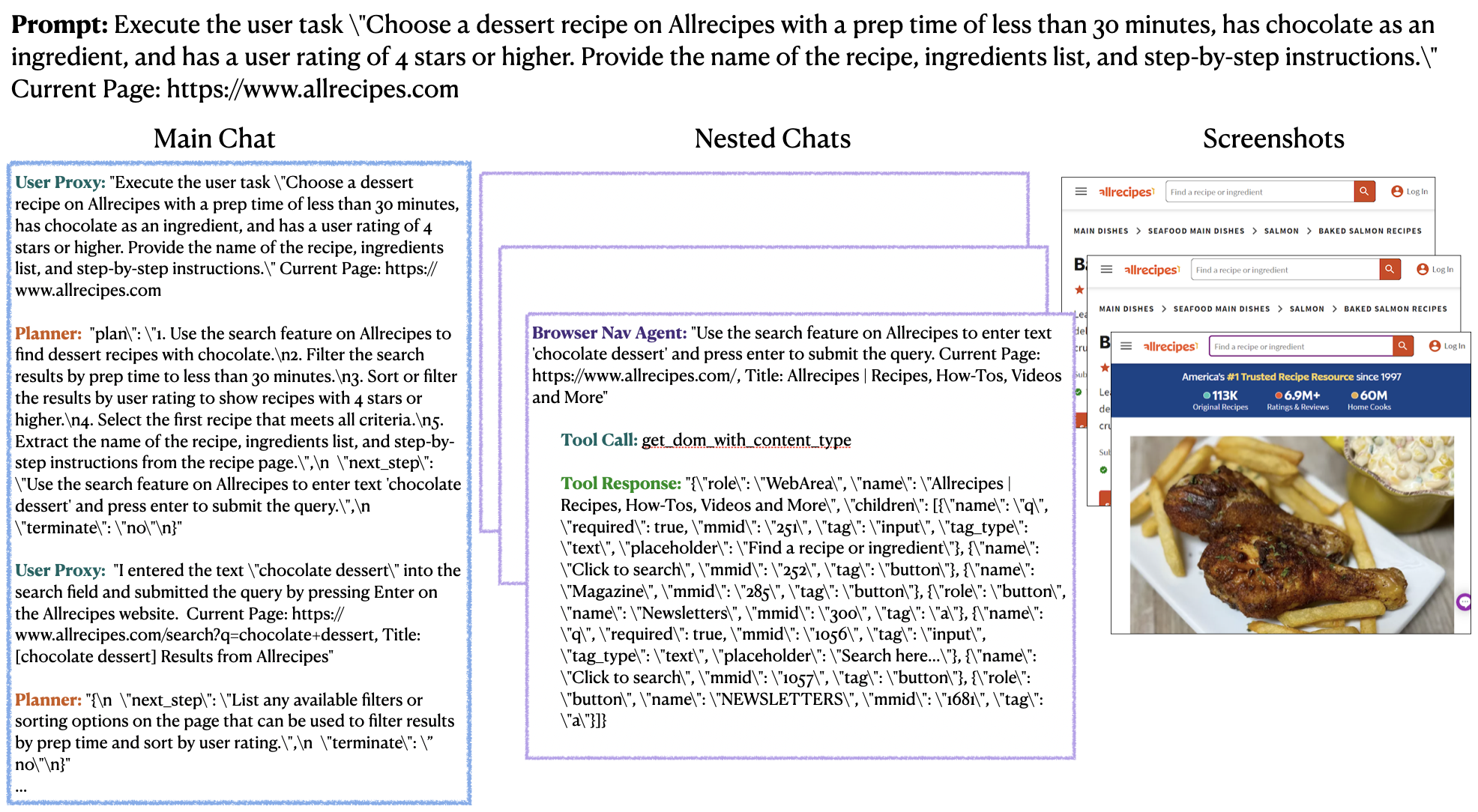}}
    \caption{Text and vision modalities from Agent-E.}
    \label{fig:agent_e_modality}
\end{figure}

We utilize Agent-E \citep{Abuelsaad2024}, a multi-agent system implemented using the AutoGen~\citep{wu2024autogen} framework. Using a hierarchy, Agent-E breaks down fine-grained navigation tasks into high-level steps and simplifies complex DOMs. The hierarchical system consists of two chats. First, a chat between the planner agent and user proxy, which plans and observes the workflow at a high level. Second, a chat between the browser agent and browser nav executor, which handles the low-level observations and browser interactions. For example, the planner agent may suggest a single-step "Search for white shoes", and the browser-agent chat would break down the actions: click the search bar, type "white shoes", and click enter. Examples of the chat logs can be seen in Figure~\ref{fig:agent_e_modality}. This hierarchical system has been shown to have state-of-the-art performance on WebVoyager task \citep{Abuelsaad2024}.

We construct our validator using three distinct modalities:
\begin{enumerate}
    \item \textbf{Task Log (Text)}: This method utilizes only the chat log between the planner agent and user proxy, containing the high-level actions and observations.
    \item \textbf{Screenshots (Vision)}: This method employs a sequence of screenshots taken throughout the workflow execution.
    \item \textbf{Screenshots (Vision) + Final Response (Multimodal)}: This method combines a sequence of screenshots with the final text response provided by the planner agent.
\end{enumerate}

In the next section, we will leverage these validation methods to improve web navigation with no additional human supervision. The implementation can be found \href{https://anonymous.4open.science/r/Agent-E-7E43}{https://anonymous.4open.science/r/Agent-E-7E43}. 

    \subsection{Self-Refine} \label{sec:self_refine_method}
\begin{figure}[h]
    \centering
    \includegraphics[width=0.7\linewidth]{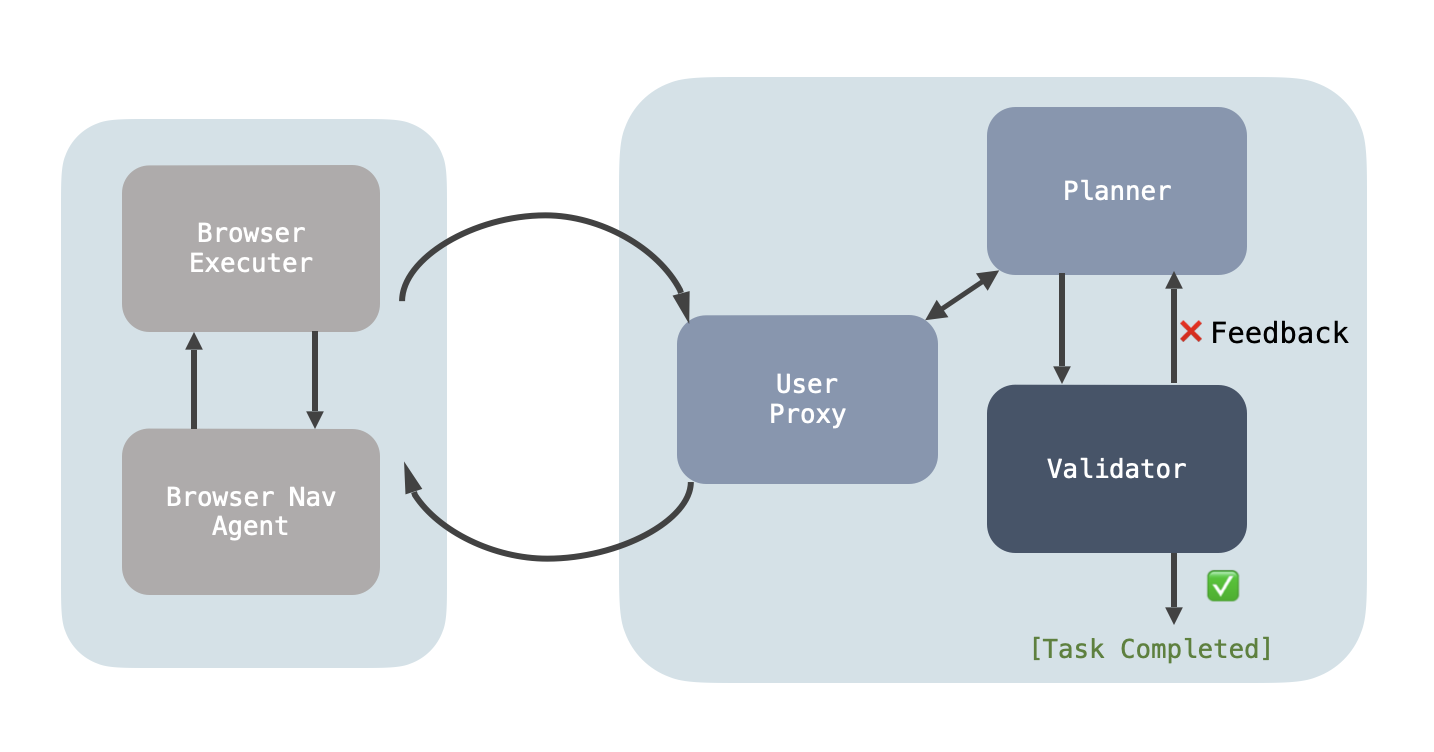}
    \caption{Multi-Agent Architecture in Agent-E with Self-Verifier.}
    \label{fig:agent_e_w/_validation}
\end{figure}

Utilizing the self-refine method \citep{Madaan2023}, our approach introduces a self-correcting mechanism for workflow failures. Our system comprises a web navigation agent that plans and executes workflows, complemented by a validator that assesses task completion and provides critical feedback. In cases where a task remains incomplete, the agent leverages the validator's feedback to revise its strategy and reattempt the task. We implement this iterative refinement process within Agent-E by integrating a verifier agent into the existing chat structure, which already includes the planner and user agent. Figure~\ref{fig:agent_e_w/_validation} provides a visual representation of the multiagent system at play.

\section{Experiments \& Results} 
    In this section, we show how self-refinement can enhance web navigation without requiring additional human supervision. First, in Section~\ref{sec:validator_result}, we demonstrate the effectiveness of our validation model. Our results highlight the characteristics of various modalities and state representations in prompting (V)LLMs-based validator models. While the validator is not perfect, we show in Section~\ref{sec:self_refine_results} that it still leads to significant performance improvements in web navigation tasks when combined with a self-refinement mechanism. Finally, in Section~\ref{sec:challenges}, we discuss key insights from our experiments to serve as valuable guidance for future practitioners.

\paragraph{Dataset} Our experiments are benchmarked using WebVoyager. This dataset comprises 643 tasks that require agents to interact with live websites. These tasks span 15 diverse websites covering various aspects of daily life, including shopping, finding news articles, and booking flights.

\paragraph{Evaluation Method} Since most tasks in the WebVoyager benchmark are open-ended and occur on live websites, there may be more than one correct workflow. A human annotator is employed to observe the agents' workflows and labels each task as `complete' or `incomplete.' A task is considered complete only if the agent successfully finishes all parts of the instructed task and remains on the designated website. The \textbf{Validator Accuracy} measures the percentage of times the validator's label and human annotator's labels match.

\paragraph{Models} We use Agent-E as our base web-navigation agent. To remain consistent with prior work benchmarks on WebVoyager \citep{he2024webvoyagerbuildingendtoendweb, Abuelsaad2024}, we utilize GPT-4-Turbo (gpt-4-turbo-preview) as a planner and browser navigator in our Agent-E implementation. For the validator, we use GPT4-Omni, GPT-4-Turbo (for text),  and GPT-4V (for vision). Our exact implementation of Agent-E and validator is available at \href{https://anonymous.4open.science/r/Agent-E-7E43}{https://anonymous.4open.science/r/Agent-E-7E43}.

\subsection{Validator Results}\label{sec:validator_result}

This section demonstrates the effectiveness of different validation models that receive different modalities of the workflow. For this experiment, we run Agent-E over each WebVoyager task. We utilize GPT4-Turbo for modalities with text only and GPT4-V for modes with vision. Then the validator and human annotators label if the task was completed by the agent.

Table~\ref{tab:validator_confusion} summarizes the performance of each modality of the validator. The Task Log (text) validator demonstrated the best performance, of $84.24\%$,  with the multimodal validator performing similarly at $83\%$. The vision validator performed notably worse than the multimodal validator, indicating the importance of the agent's final answer in some tasks. However, between $17.68-19.67\%$ of tasks were labeled True Negatives. This indicates there are cases where LLMs cannot correctly plan a web navigation task, but can determine that a workflow is not completed successfully. 

\begin{table}[h!]
\centering
\begin{adjustbox}{width=0.8\linewidth,center}
\renewcommand{\arraystretch}{1.2}
\begin{tabular}{lccccc} 
\toprule
\textbf{} 
    & \textbf{\thead{True \\ Positive}}  
    & \textbf{\thead{True \\ Negative}}  
    & \textbf{\thead{False \\ Positive}}  
    & \textbf{\thead{False \\ Negative}}  
    & \textbf{\thead{Validator \\ Accuracy}} \\
\midrule
Task Log (text) & 66.56\% & 17.68\% & 7.40\% & 8.36\% & \textbf{84.24\%} \\
Screenshots (vision)  & 52.03\% & 18.02\% & 3.60\% & 26.13\% & 70.04\% \\ 
Screenshot + Final Response (multimodal) & 63.33\% & 19.67\% & 5.00\% & 12.00\% & 83.00\% \\
\bottomrule
\end{tabular}
\end{adjustbox}
\caption{Confusion matrix and accuracies of validators.}
\label{tab:validator_confusion}
\end{table}

The task-specific performance of the validators can be seen in Table~\ref{tab:validator_sites}. Although overall the text validator outperforms the vision validator, this section indicates there are tasks where the vision validator performs better. The Booking.com site has a notably difficult DOM, making it consistently one of the most challenging tasks for web navigation. For these tasks, the vision validator significantly outperforms the Task Log (text) validator. Additionally, the vision validator also performed notably better for Google Flight tasks. This website is highly dynamic and requires navigating widgets which are difficult to represent in the DOM. On the other hand, highly text-based tasks perform significantly better with some text modality (e.g., Google, Huggingface, Wolfram Alpha). This difference in task-specific performance highlights the benefit of having task-specific validators.

\begin{table}[h!]
\centering
\begin{adjustbox}{width=\linewidth,center}
\renewcommand{\arraystretch}{1.2}
\begin{tabular}{lccccccccc}
\hline
\textbf{} & \textbf{Amazon} & \textbf{Allrecipes} & \textbf{Apple} & \textbf{Arxiv} & \textbf{\thead{BBC \\News}} & \textbf{Booking.com} & \textbf{Coursera} & \textbf{\thead{Cambridge \\Dictionary}} \\
\hline
Task Log (text) & 75.61\% & 82.22\% & 79.07\% & 88.37\% & \textbf{92.86\%} & 69.77\% & \textbf{95.24\%} & 93.02\% \\
Screenshots (vision) & \textbf{77.50\%} & 80.00\% & 80.49\% & 88.37\% & 90.48\% & \textbf{85.00\%} & 92.86\% & 95.24\% \\
Screenshot + Final Response (multimodal) & 65.85\% & \textbf{84.44\%} & \textbf{81.40\%} & \textbf{90.70\%} & 88.10\% & 83.72\% & 92.86\% & \textbf{95.35\%} \\
\hline
\textbf{} & \textbf{ESPN} & \textbf{Github} & \textbf{Google} & \textbf{\thead{Google \\Maps}} & \textbf{\thead{Google \\Flights}} & \textbf{\thead{Huggingface}} & \textbf{\thead{Wolfram \\Alpha}} & \textbf{Overall} \\
\hline
\textbf{Task Log (text)} & 93.18\% & 90.24\% & \textbf{93.02\%} & \textbf{90.24\%} & 90.48\% & 70.00\% & \textbf{91.30\%} & \textbf{84.24}\% \\
\textbf{Screenshots (vision)} & \textbf{95.35\%} & \textbf{94.74\%} & 65.38\% & 66.67\% & 92.11\% & 56.25\% & 33.33\% & 70.04\% \\
\textbf{Screenshot + Final Response (multimodal)} & \textbf{95.35\%} & 87.80\% & 92.31\% & 90.48\% & \textbf{95.12\%} & 75.00\% & 90.48\% & 83.00\% \\
\hline
\end{tabular}
\end{adjustbox}
\caption{Validator accuracy by website.}
\label{tab:validator_sites}
\end{table}

\subsection{Self-Refinement Results}\label{sec:self_refine_results}
In Section~\ref{sec:validator_result}, we demonstrate that our validators can, in some cases, detect failed workflows. In this section, we leverage these validators using a self-refinement approach. In this experiment, Agent-E completes and revises its workflows based on feedback from our validator. The experiment is conducted on 322 tasks from the WebVoyager dataset, with the selected tasks evenly distributed across the websites — specifically, the task\_id with an even number in the WebVoyager dataset.

Table~\ref{table:webVoyager_accuracies} compares the performance of Agent-E with and without Self-Validation across text and vision tasks, using GPT-4 Omni as the validation model. The greatest performance improvement is observed in vision tasks, where Self-Validation increases accuracy from 76.2\% to 81.24\%. This boost in performance highlights the utility of our validator in enhancing web navigation.

\begin{table}[h!]
\centering
\renewcommand{\arraystretch}{1.2}
\begin{adjustbox}{width=0.48\linewidth}
\begin{tabular}{lccccc}
\hline
 & \textbf{\thead{Success \\ Rate}}\\ \hline
\textbf{Self-Refine (Text)} & 79.81\%\\ 
\textbf{Self-Refine (Vision)} & \textbf{81.24\%}\\ 
\textbf{Self-Refine (Vision + Final Text Response)} & 79.35\% \\ 
\textbf{Agent-E (text)} & 76.2\% \\
\hline
\end{tabular}
\end{adjustbox}
\caption{The success rate of web-navigation methods on 322 WebVoyager tasks.}
\label{table:webVoyager_accuracies}
\end{table}

The performance of each method, by website, can be seen in Table~\ref{table:model_eval_by_task}. Although the performance of different modalities is similar — with performance ranging from $79.25\%-81.24\%$ — there are notable differences depending on the task. Tasks that are primarily text-based and performed on simple websites tend to perform best with the text validator (e.g., Google Search, Arxiv, Hugging Face, WolframAlpha). In contrast, websites that are highly dynamic with complex DOMs perform better with the vision validator (e.g., Google Flights and Booking.com). Notably, Booking.com shows performance gains of over 13\% using vision over text. Using a hierarchy of agents makes DOMs more interpretable for an LLM, the vision modality proves to be more effective when the site is sufficiently complex. Additionally, we found that the multimodal validator does not outperform the text-only or vision-only validators, suggesting the potential need for specialized, task-specific validators. It is worth noting that our multimodal validator only contains text from the final planner response.

\begin{table}[h!]
\centering
\begin{adjustbox}{width=\linewidth,center}
\renewcommand{\arraystretch}{1.2}
\begin{tabular}{lccccccccc}
\hline
\textbf{} & \textbf{Amazon} & \textbf{Allrecipes} & \textbf{Apple} & \textbf{Arxiv} & \textbf{\thead{BBC \\News}} & \textbf{Booking.com} & \textbf{Coursera} & \textbf{\thead{Cambridge \\Dictionary}} \\
\hline
\textbf{Task Log (text)} & \textbf{89.47}\% & 78.26\% & \textbf{100.00\%} & \textbf{90.48}\% & \textbf{95.00}\% & 23.81\% & 80.95\% & 86.36\% \\ 
\textbf{Screenshots (vision)}  & 73.68\% & \textbf{91.30\%} & 86.36\% & 66.67\% & 80.95\% & \textbf{41.18}\% & \textbf{95.24\%} & 95.45\% \\
\textbf{Screenshot + Final Response (multimodal)} & 80.00\% & \textbf{91.30\%} & 90.91\% & 57.14\% & 80.95\% & 22.73\% & 90.48\% & \textbf{95.45\%} \\ 
\hline
\textbf{} & \textbf{ESPN} & \textbf{Github} & \textbf{Google} & \textbf{\thead{Google \\Maps}} & \textbf{\thead{Google \\Flights}} & \textbf{\thead{Huggingface}} & \textbf{\thead{Wolfram \\Alpha}} & \textbf{Overall} \\
\hline
\textbf{Task Log (text)} & 72.73\% & 85.00\% & \textbf{95.24\%} & 76.19\% & 61.90\% & 70.00\% & \textbf{91.30\%} & 79.81\% \\
\textbf{Screenshots (vision)}  & \textbf{100.00\%} & \textbf{100.00\%} & 76.19\% & 76.19\% & \textbf{85.71\%} & 63.18\% & 77.27\% & \textbf{81.24\%} \\
\textbf{Screenshot + Final Response (multimodal)} & 95.45\% & \textbf{100.00\%} & 95.24\% & \textbf{80.95\%} & 61.90\% & \textbf{85.71\%} & 60.00\% & 79.35\% \\ 
\hline
\end{tabular}
\end{adjustbox}
\caption{The success rate of web-navigation methods on 322 WebVoyager tasks by website.}
\label{table:model_eval_by_task}
\end{table}

\subsection{Technical Challenges and Observations}\label{sec:challenges}
During our experimentation, we encountered several technical challenges and made noteworthy observations:

\paragraph{Screenshot Acquisition} Due to variability in website loading times and browser performance, screenshot capture for sequence information was often unsuccessful. To mitigate this issue, we implemented multiple screenshot attempts. Yet, in some cases, all attempts failed. When presenting results for the vision validator, we omitted tasks where no screenshots were captured throughout the workflow.

\paragraph{JSON Formatting} Our validator is prompted to output two parts: 1) the task evaluation and 2) the feedback. To ensure reliable parsing of these two parts, we required the agent to output them in JSON format. In instances where the JSON formatting failed, we had the validator label the workflow as incomplete and proceed with the refinement process.

\begin{figure}[htbp]
    \centering
    \begin{tcolorbox}[
        colback=white,
        colframe=black,
        boxrule=1pt,
        arc=4pt,
        left=10pt,
        right=10pt,
        top=10pt,
        bottom=10pt,
        width=\textwidth
    ]
    \ttfamily
    \begin{minipage}{\dimexpr\linewidth-20pt\relax}
        \tiny
        \textbf{Task:} Reserve a hotel in downtown Chicago with a rating of 9 or higher for a stay from September 20-27, 2024, which offers free cancellation and includes a fitness center.

        \vspace{0.5em}

        \textbf{Validator Reason:} The workflow was successfully completed as all the criteria were met: a hotel in downtown Chicago with a rating of 9 or higher was found, it offers free cancellation, includes a fitness center, and the stay is from September 20-27, 2024.
    \end{minipage}
    
    \vspace{1em}
    
    \begin{minipage}{\dimexpr\linewidth-20pt\relax}
        \centering
        \includegraphics[width=0.9\textwidth]{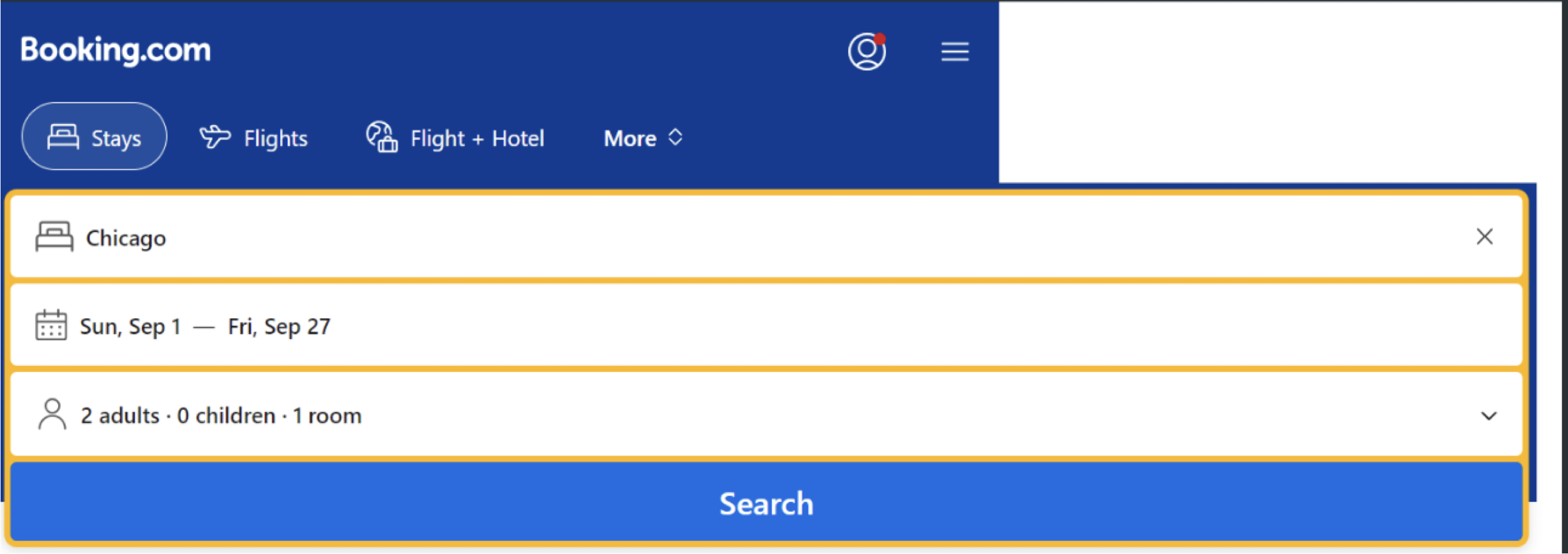}
    \end{minipage}
    \end{tcolorbox}
    \caption{Task log validator failure case.}
    \label{fig:booking_failure}
\end{figure}

\paragraph{Modality Specific Performance} The combined vision and text approach demonstrated comparable performance to text-only methods and surpassed vision-only approaches in most tasks. We made two notable observations in cases which led to such performance:

\begin{itemize}
    \item For question-answering tasks, the model can more easily verify answers when explicitly stated in text rather than inferred from a series of screenshots. This explains the large jump in performance between the screenshot and multimodal validator for tasks like Google. 
    
    \item The vision-based models exhibited superior performance on complex websites such as booking.com and flights.com. Oftentimes, the navigator would unsuccessfully utilize filters and widgets on such websites, yet the planner agent would assume that the task was completed successfully. In such cases, the task log verifier was seen to be overly optimistic about the task completion. An example can be seen in Figure~\ref{fig:booking_failure}.
\end{itemize}

 The variability in performance across different modalities and websites underscores the importance of implementing robust, multi-modal validation strategies. Future implementations can consider adaptive approaches that can dynamically select the most appropriate validation method based on the task and website characteristics. Additionally, practitioners should be prepared to handle technical issues such as inconsistent screenshot capture and JSON parsing failures, potentially by implementing fallback mechanisms or alternative data collection methods.

\section{Conclusion} 
    In this study, we developed an effective validator and integrated it into a self-refinement mechanism. This approach allowed web agents to detect and self-correct workflow errors without additional human supervision. Our experiments on the WebVoyager benchmark demonstrated significant improvements, boosting task completion rates from 76.20\% to 81.24\%, surpassing prior state-of-the-art performance. While the overall performance across modalities was comparable, with success rates ranging from 79.25\% to 81.24\%, we observed significant task-specific variations in effectiveness. This highlights that the best modality for validation is task-dependent, suggesting the need for specialized validators.

While we encountered technical challenges, particularly in screenshot acquisition, our findings underscore the complexity of real-world web environments and provide valuable insights for future research. This work contributes to the development of more reliable and adaptable web agents. It brings us closer to the goal of creating autonomous digital assistants capable of navigating the intricacies of the modern web.

\bibliographystyle{plainnat}
\bibliography{references}

\appendix

\end{document}